\documentclass[runningheads]{llncs}
\usepackage[T1]{fontenc}
\usepackage{graphicx}
\usepackage{color}
\usepackage{url}

\usepackage{booktabs}

\usepackage{graphicx}  
\usepackage{float}  
\usepackage{subfigure}  

\usepackage{multirow} 
\usepackage{colortbl} 
 \usepackage{makecell} 

\usepackage{caption}

\usepackage{CJKutf8}

\usepackage[marginal]{footmisc}

\newcommand \footnoteONLYtext[1]
{
	\let \mybackup \thefootnote
	\let \thefootnote \relax
	\footnotetext{#1}
	\let \thefootnote \mybackup
	\let \mybackup \imareallyundefinedcommand
}

\usepackage{bbding}
\usepackage{hyperref}

\begin{document}

\title{FAIR: A Causal Framework for Accurately Inferring Judgments Reversals}
\titlerunning{FAIR}
%

\author{Minghua He\inst{1} \and
Nanfei Gu\inst{2} \and
Yuntao Shi\inst{1} \and
Qionghui Zhang\inst{3} \and
Yaying Chen\inst{1(}\Envelope\inst{)}
}
\authorrunning{M. He et al.}
%

\institute{College of Computer Science and Technology, Jilin University, Changchun, China \and
School of Law, Jilin University, Changchun, China \and
Gould School of Law, University of Southern California, Los Angeles, USA
}

\maketitle              

\footnoteONLYtext{First Author and Second Author contribute equally to this work.}

\begin{abstract}
Artificial intelligence researchers have made significant advances in legal intelligence in recent years. However, the existing studies have not focused on the important value embedded in judgments reversals, which limits the improvement of the efficiency of legal intelligence. In this paper, we propose a causal \textbf{F}ramework for \textbf{A}ccurately \textbf{I}nferring case \textbf{R}eversals (FAIR), which models the problem of judgments reversals based on real Chinese judgments. We mine the causes of judgments reversals by causal inference methods and inject the obtained causal relationships into the neural network as a priori knowledge. And then, our framework is validated on a challenging dataset as a legal judgment prediction task. The experimental results show that our framework can tap the most critical factors in judgments reversal, and the obtained causal relationships can effectively improve the neural network's performance. In addition, we discuss the generalization ability of large language models for legal intelligence tasks using ChatGPT as an example. Our experiment has found that the generalization ability of large language models still has defects, and mining causal relationships can effectively improve the accuracy and explain ability of model predictions.

\end{abstract}

\section{Introduction}
Legal intelligence is dedicated to assist legal tasks through the application of artificial intelligence. Data resources in the legal field are mainly presented in the form of textual documents, and China has the world's largest database of judgment documents, which can be further explored for its significant value through natural language processing(NLP). In recent years, with the increase of computing power and data scale, deep learning algorithms have developed rapidly and gradually become the mainstream technology of legal intelligence. ChatGPT is a typical large language model(LLM) that has triggered intense discussions, and its generalization ability in the legal field also needs to be studied.

Artificial intelligence researchers have put forth many fruitful efforts in advancing the use of deep learning in legal intelligence. Several works in recent years have contributed very rich legal data resources to the natural language processing community~\cite{yao2022leven,chen2020joint,xiao2018cail2018}, and these datasets together form the basis of legal intelligence research. Based on these datasets, researchers have designed diverse legal AI tasks based on the practical needs of the legal domain, among which representative tasks include legal judgment prediction (LJP)~\cite{csulea2017predicting}, legal case matching~\cite{yu2022explainable}, legal entity extraction~\cite{chen2020joint}, etc. Based on natural language processing techniques, researchers have developed corresponding solutions for these tasks and applied them in judicial practice.

However, the established work neglects the issue of judgments reversals, which is the area most closely linked to the application of law. According to our statistics, the percentage of revision of judgments reaches 14.63\% of all judgments in China, which is a non-negligible part. The problem of judgments reversals is directly related to the direction of application of AI techniques and the effect of models. In the LJP task, extracting the causal relationship in judgments reversals as a priori knowledge helps to improve the accuracy as well as interpretability of model prediction.

Although the problem of judgments reversals has important theoretical and practical value, there are major challenges in the research. 1) It is more difficult to model the actual situation of reversals of judgments with high quality. The difficulty of this part of the work is that it is difficult to uncover all the factors that influence the judgment, and it is difficult to quantify and analyze factors such as judges' subjective will. 2) It is difficult to directly apply the prior knowledge to the improvement of neural networks. How to make neural networks efficiently use prior knowledge from different domains has been one of the challenges of research in artificial intelligence.

In this paper, we propose a causal \textbf{F}ramework for \textbf{A}ccurately \textbf{I}nferring judgments \textbf{R}eversals (FAIR), which mines why revisions occur based on causal inference, which is the process of exploring how one variable $T$ affects another variable $Y$. In the construction of FAIR, first, the causal graph is initially modeled with the help of legal experts by training an encoder to remove the redundant constraints in the graph. Then, the causal effects between different variables are estimated quantitatively using a causal inference algorithm. Finally, the obtained causal knowledge is injected into the neural network model of the downstream task, which can effectively improve the performance of the model.

While the recent rise of Large Language Models (LLMs) has had a huge impact on the natural language processing community, we are also interested in the generalization ability of LLMs on legal intelligence tasks. We design challenging experiments to explore the knowledge exploitation ability and reasoning power of LLMs in the legal domain, and add LLMs as comparisons in the evaluation experiments of the FAIR framework. The experiments reveal some current limitations of LLM and demonstrate that the generalization ability of LLM can be enhanced by causal knowledge mining and injection.

Our main contributions are as follows: 1) We propose FAIR, a causal Framework for Accurately Inferring judgments Reversals, and better mine the causal relationships in complex legal judgments based on causal inference to uncover the reasons for judgments reversals. 2)The results obtained from performing the LJP task on a real legal dataset indicate that it is effective to improve the performance of neural networks by injecting prior knowledge. 3) We explore the knowledge utilization capability and inference capability of LLM in the legal domain. By comparing our framework with LLM, we reveal some limitations of LLM currently existing and proposed ways to improve its generalization ability.

\section{Related Work}

\subsection{Legal Intelligence}
  Legal Intelligence focuses on applying natural language processing techniques to the legal domain, for which researchers have designed diverse tasks and provided rich data resources. CAIL2018~\cite{xiao2018cail2018} is a large-scale Chinese legal dataset designed for the LJP task, focusing on LJP in the criminal law domain. LEVEN~\cite{yao2022leven} considers the legal event detection task. FSCS~\cite{niklaus2021swiss} provides multilingual data for the LJP task and studies the legal differences in different regions. LeSICiN~\cite{paul2022lesicin} designed the law and regulation identification task, using graphs to model the citation network between case documents and legal texts. MSJudge~\cite{ma2021legal} describes a courtroom argument scenario with multi-actor dialogues for the LJP task. Some work has attempted to provide solutions to the above tasks using natural language processing techniques, and Lawformer~\cite{xiao2021lawformer} has designed a pre-training model for legal text training. EPM~\cite{feng2022legal} considers implicit constraints between events in the LJP task. NSCL~\cite{gan2022exploiting} attempts to use contrast learning to capture the subtle differences between legal texts in the LJP task. QAjudge~\cite{zhong2020iteratively} uses reinforcement learning to provide interpretable predictions for LJP. However, these works have not taken into account the issue of judgments reversals, which is directly related to the application of the law.

\subsection{Causal Inference for Legal Domain}
  Recent work has attempted to use causal inference to provide more reliable explanations and greater robustness for legal intelligence. Liepina~\cite{liepina2018causal} introduces a semi-formal causal inference framework to model factual causality arguments in legal cases. Chockler~\cite{chockler2015causal} investigates the problem of legal attribution of responsibility using causal inference to capture complex causal relationships between multiple parties and events. GCI~\cite{liu2021everything} designs a causal inference framework for unlabeled legal texts, using a graph-based approach to construct causal graphs from factual descriptions. Evan~\cite{iatrou2022normative} uses causal inference to provide explanations for binary algorithms in legal practice. Law-Match~\cite{sun2022law} considers the influence of legal provisions in legal case-matching tasks and incorporates them as instrumental variables in causal graphs. Chen et al~\cite{chen2022knowledge} investigated the problem of pseudo-correlation error introduced by pre-trained models and eliminated this error by learning the underlying causal knowledge in legal texts.

\section{Methodology}
\begin{figure}
\includegraphics[width=\textwidth]{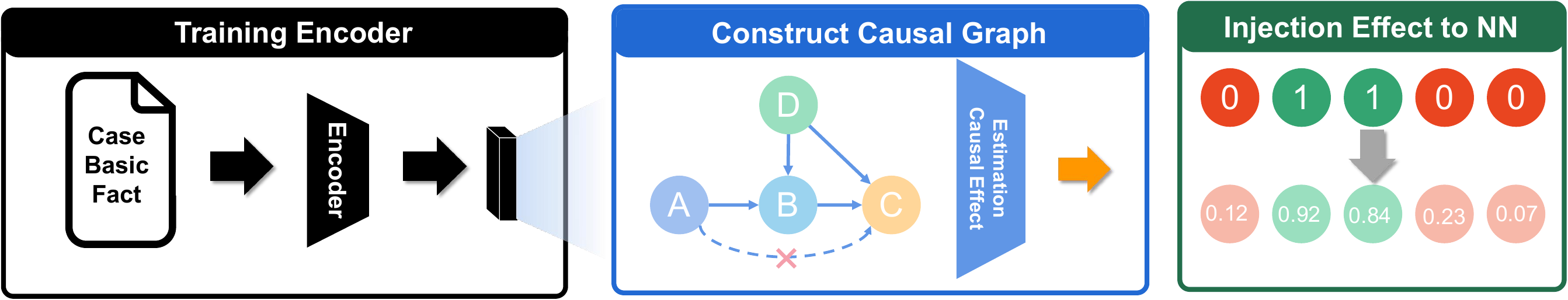}
\caption{Overall structure of FAIR} \label{fair}
\end{figure}

Our framework FAIR consists of three main parts, including causal graph modeling, estimating causal effects on the modeled causal graph, and injecting causal effects into the neural network. Figure \ref{fair} illustrates the structure of FAIR.
\vspace{-8ex}

\subsection{Modeling Causal Graph}

\subsubsection{Preliminary Modeling and Analysis}
\begin{figure}[H]
	\centering  
	\subfigbottomskip=2pt 
	\subfigcapskip=5pt 
	\subfigure[Preliminary Causal Graph]{
		\includegraphics[width=0.45\linewidth]{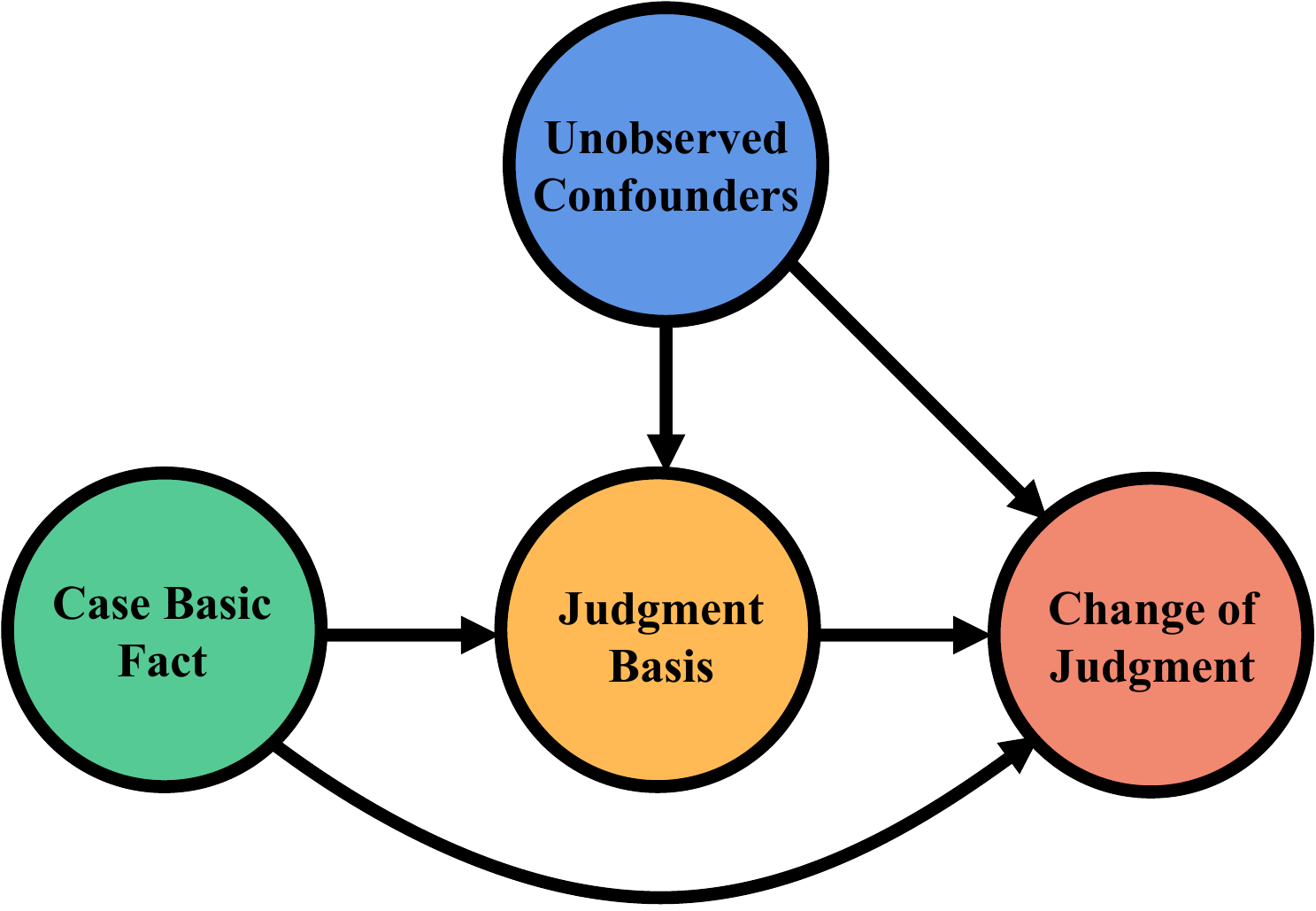}\label{cause1}}
	\subfigure[Target Causal Graph]{
		\includegraphics[width=0.45\linewidth]{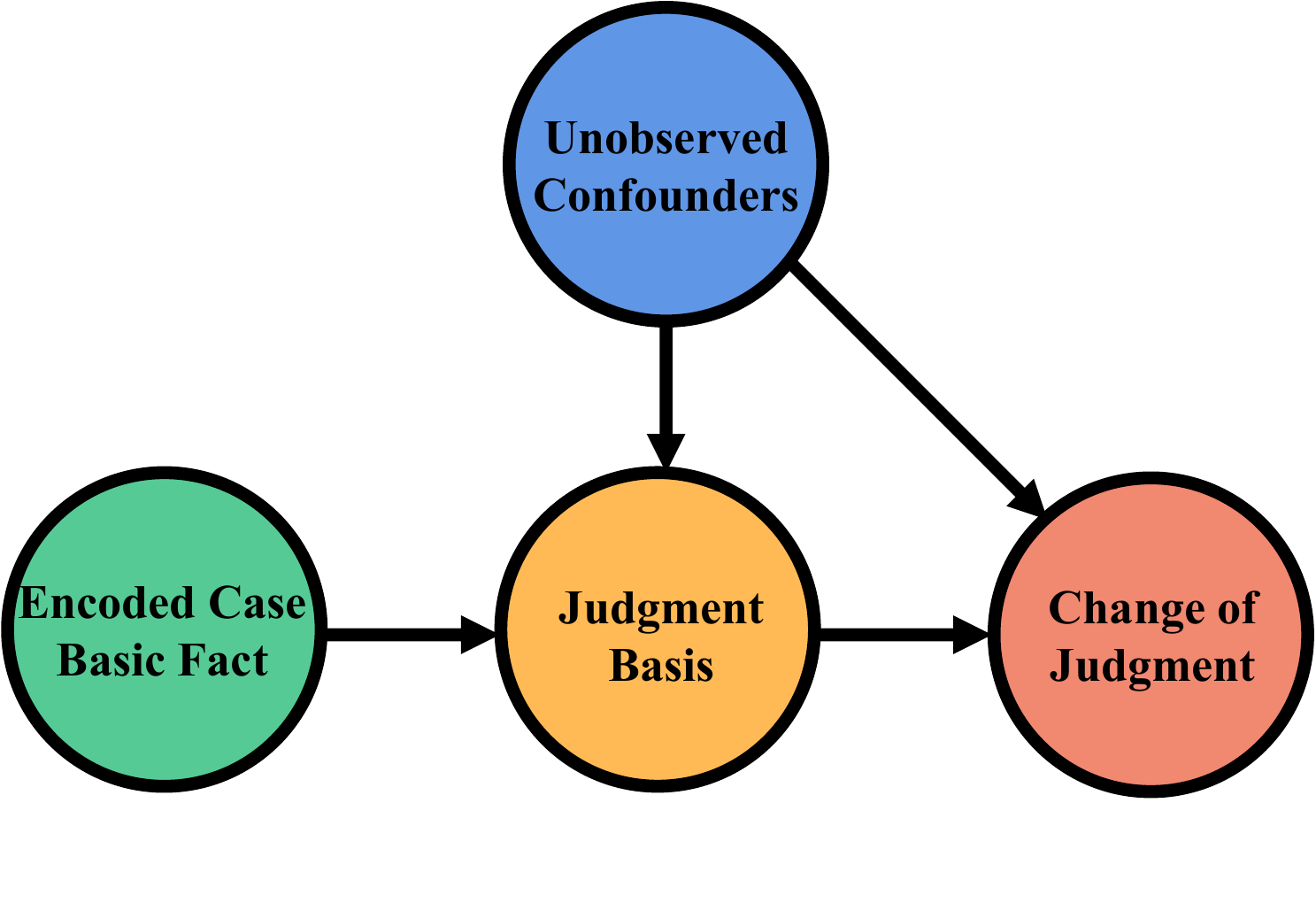}\label{cause2}}
  \caption{Preliminary and Target Causal Graph} 
\end{figure}

Before conducting a quantitative analysis of causal effects, we need to model the problem based on prior knowledge to ensure the clarity of causal assumptions, and the modeling results are given in the form of a causal graph. We describe the possible causal relationships in the judgment with the help of legal experts as Figure \ref{cause1}. However, in Figure \ref{cause1}, we cannot directly estimate the causal relationship between "Judgment Basis" and "Case Basic Fact" because there are multiple causal paths between them, and we need to block the paths that are not directly connected. Considering the presence of unobserved confounders in Figure \ref{cause1}, we choose the instrumental variable method to block the paths through the confounders, which means that "Case Basic Fact" will be used as an instrumental variable, and it needs to satisfy the correlation and exogeneity. To ensure exogeneity, we need to block the direct path from the instrumental variable to the outcome, which means we need to extract the part of the instrumental variable that is relevant to the treatment and not relevant to the outcome, and we do this using a law article prediction task.

 \vspace{-8ex}

\subsubsection{Task Definition}
Given a factual description of the judgment containing $n$ tokens $X = \{x1, x2, ... , x_n\}$ and a set $L = \{l1, l2,..., lm\}$ containing $m$ legal entries, we want the model to find a many-to-one mapping $F$ from set $X$ to a subset of $L$, and the result of the mapping is denoted as an $m$-dimensional multi-hot vector. This task can be understood as a multi-label classification task.
\subsubsection{Encoder}
We use the pre-trained model Lawformer as an encoder and fine-tune it in the law article prediction task as a way to capture the features we need. First, we use Lawformer to encode $X = \{x1, x2, ..., x_n\}$.
\begin{equation}
H = Encoder(x_i)
\end{equation}
Then, the encoded representation $H$ is fed into a linear layer and the dimension of $Output$ is $m$, as same as the number of labels.
\begin{equation}
Output = \{out_1, out_2,...out_m\} =Linear(H)
\end{equation}
Considering the possible data imbalance of the real labels, we use ZLPR~\cite{su2022zlpr} as the loss function.
\begin{equation}
L_{zlpr} = log(1 + \sum\limits_{i\in\Omega_{pos}}e^{-out_i}) + log(1 + \sum\limits_{j\in\Omega_{neg}}e^{out_j})
\end{equation}
where $\Omega_{pos}$ is the set of positive samples and $\Omega_{neg}$ is the set of negative samples. After extracting the features by Encoder, Figure \ref{cause2} shows the causal graph we finally obtained.

\subsection{Causal Effects Estimation}
The estimation of causal effects requires controlling for confounders to ensure the accuracy of the results, which we discuss in detail in Section 3.1, where we use "Encoded Case Basic Fact" as an instrumental variable to ensure this.

We use Average Treatment Effect (ATE) as a quantitative criterion for the causal effect. Suppose $T$ is the intervention variable, $ p(Y|do(T = a))$ is the interventional distribution, and $Y$ is the target variable. Then, under the reference condition $ T= b $, the $ATE$ after imposing the intervention $ T= a $ is described as
\begin{equation}
ATE(a,b)=E_p(Y|do(T=a))-E_p(Y|do(T=b))
\end{equation}
Under the condition that instrumental variables are used, the computation of $ATE$ can be described in the following form. We let $U$ be the confounder and $Z$ be the instrumental variable, and suppose that $Y = \delta T + \alpha U$, we have $YZ = \delta TZ + \alpha UZ$, and since $Z$ is not affected by $U$, the above equation is equivalent to $YZ = \delta TZ$, then the causal estimator $\delta = YZ * (TZ) ^{-1}$, it is easy to find that $\delta$ is exactly the unbiased estimate of $ATE$, i.e.
\begin{equation}
ATE(a,b) = \frac{E(Y|Z = a)-E(Y|Z = b)}{E(T|Z = a)-E(T|Z = b)}
\end{equation}

\subsection{Causal Smoothing}
In this subsection, we propose a method called Causal Smoothing to inject the causal effects estimated by FAIR into a neural network. We draw inspiration from the widely used Label Smoothing~\cite{szegedy2016rethinking}. If $y_i$ is the label of the classification task, $y_i$ is $0$ or $1$ in the hard label case. Label Smoothing replaces the hard label $y_i$ with a soft label
\begin{equation}
p_i = (1 - \epsilon)y_i + \frac{\epsilon}{K}
\end{equation}
where $K$ is the number of categories of labels and $\epsilon$ is the hyperparameter, which is the same for all samples in training. Causal Smoothing modifies $\epsilon$ to $\epsilon_i = \omega\sum\limits_{j = 1}^{m} ATE(t_{ij},0)$, where $t _{ij}$ is the value of the $j$th treatment in the $i$th sample, $m$ is the number of treatments, and $\omega$ is the hyperparameter. In Causal Smoothing, the soft label can be expressed as
\begin{equation}
p_{i+causal} = (1 - \omega\sum\limits_{j = 1}^{m} ATE(t_{ij},0))y_i + \frac{\omega\sum\limits_{j = 1}^{m} ATE(t_{ij},0)}{K}
\end{equation}

\section{Experiments and Evaluation}
In this section, we apply FAIR to a specific legal scenario and test the inference results of the framework in a downstream task of legal intelligence. We have chosen the determination of labor relationship for over-aged labors as the legal issue for the experiment, and the legal judgment prediction task as the downstream task. Our experimental results have shown the superiority of FAIR in this context.
\subsection{Dataset}
Currently, published datasets do not consider our research topic of judgments reversals and are too coarse-grained to meet the needs of fine-grained tasks. They do not differentiate between initial and appellate judgments in legal cases, and we cannot obtain the required labels for FAIR. Therefore, we construct a dataset of unstructured judgments from the internet and used regular expressions to extract the labels for our experiment. We used this method because Chinese judgments have a relatively fixed structure, and it is fairly accurate for us to extract the required labels. We download and extract all of the judgments on the determination of labor relationships for over-aged labors issues from the pkulaw\footnote[1]{\url{https://www.pkulaw.com}} website. These judgments are real and challenging, as they involve complex legal issues and difficult factual determinations. The number of training sets is 5785, the number of validation sets is 883, and the number of test sets is 416. We choose the challenging second trial data as the test set, in which the number of judgments reversals is 98.

\subsection{Experimental Setup}

\subsubsection{Causal Acquisition} 
We first design a law article prediction task to train the feature encoder, and selected the most important four law articles related to the determination of labor relationship for over-aged labors as the labels. The encoder was initialized by the pre-trained model Lawformer, and ZLPR was used as the loss function during training. We use the encoded inputs as instrumental variables to construct causal graphs, using the instrumental variables approach provided by the dowhy~\cite{sharma2021dowhy} framework for inference.

\subsubsection{Legal Judgement Prediction} 
In the legal field, LJP task requires the model to predict the outcome of a decision based on the basic fact of the input case. We chose the mainstream LJP models as the baseline, including Lawformer, Longformer, Bert~\cite{devlin2018bert}, and Bi-LSTM, which were trained under the condition of no causal knowledge and injected causal knowledge respectively. Since Bert only accepts inputs up to 512 tokens in length, we adopt a truncated input and max-pooling approach to obtain the input for the classification layer.

\subsubsection{Causal Knowledge Injection} 
We used Causal Smoothing, introduced in Section 3, to inject causal knowledge into the model, and the hyperparameter $\omega$ was set to 0.1 for the experiments. We used Label Smoothing as a control, and the hyperparameter $\epsilon$ was set to 0.1. in addition to the mainstream models described above, we also explored the performance of large language models for the LJP task, and we chose OpenAI's ChatGPT as the experimental subject and provided it with a priori causal knowledge through different prompts. All the above experiments use Adam as the optimizer, and the Learning rate is set to 1e-4.

\subsection{Main Result}
Table \ref{result} shows the experimental results of FAIR on the mainstream model of the LJP task. From it, we find that the causal knowledge obtained by inference of FAIR improves all baselines, with significant improvements on both F1 values and Acc. Specifically, for F1 values, the improvements on baselines reach 1.92, 4.88, 1.02, and 12.82, respectively; for Acc, the improvements on baselines reach 4.81, 0.96, 11.54, and 2.41, respectively.

The Bi-LSTM model has the most significant improvement in F1 values, which we believe is because the Bi-LSTM model is not capable of capturing features for long texts, and it is difficult to learn effective knowledge during the training process, so the injection of causal knowledge is a very significant improvement for Bi-LSTM. the Bert model has the largest improvement in Acc, which we believe is because the transformer model can make good use of causal knowledge. Lawformer has the best overall performance without injecting causal knowledge, which proves the advantage of legal text pre-training. In addition, we can find that LLM still has some gaps with supervised learning models in downstream tasks of legal intelligence, which we discuss in detail in Section 6.

\begin{table}
    \caption{Experimental results of FAIR on LJP task.}\label{result}
\begin{center}
\resizebox{\textwidth}{!}{
\setlength{\tabcolsep}{5mm}{
\begin{tabular}{ccccc}
   \toprule
   \textbf{Models} & \textbf{P} & \textbf{R} & \textbf{F1} & \textbf{Acc}\\
   \toprule
    Lawformer & 48.97 & \textbf{65.75} & 56.14 & 63.94\\
    $\rm Lawformer+Causal$ & \textbf{54.87} & 61.64 & \textbf{58.06} $(\uparrow1.92)$ & \textbf{68.75} $(\uparrow4.81)$ \\
    \midrule
    Longformer & 44.44 & 71.23 & 54.73 & 58.65\\
    $\rm Longformer+Causal$ & \textbf{45.92} & \textbf{84.93} & \textbf{59.61} $(\uparrow4.88)$ & \textbf{59.61} $(\uparrow0.96)$\\
    \midrule
    Bert & 46.03 & \textbf{79.45} & 58.29 & 60.09\\
    $\rm Bert+Causal$ & \textbf{59.72} & 58.90 & \textbf{59.31} $(\uparrow1.02)$& \textbf{71.63} $(\uparrow11.54)$\\
    \midrule
    Bi-LSTM & 38.70 & 16.43 & 23.07 & 61.53\\
    $\rm Bi-LSTM+Causal$ & \textbf{47.72} & \textbf{28.76} & \textbf{35.89}$(\uparrow12.82)$ & \textbf{63.94}$(\uparrow2.41)$\\
    \midrule
    ChatGPT & 39.69 & \textbf{71.23} & \textbf{50.98} & 51.92\\
    $\rm ChatGPT+Prior$ & 39.44 & 58.90 & 47.25 & 53.84\\
    $\rm ChatGPT+Prior^*$ & \textbf{41.23} & 54.79 & 47.05 & \textbf{56.73}$(\uparrow4.81)$\\
   \bottomrule
\end{tabular}
}
}
\end{center}
\end{table}

\section{Analysis}

\subsection{Robustness of Inference Results}
\begin{figure}[H]
\includegraphics[width=1\textwidth]{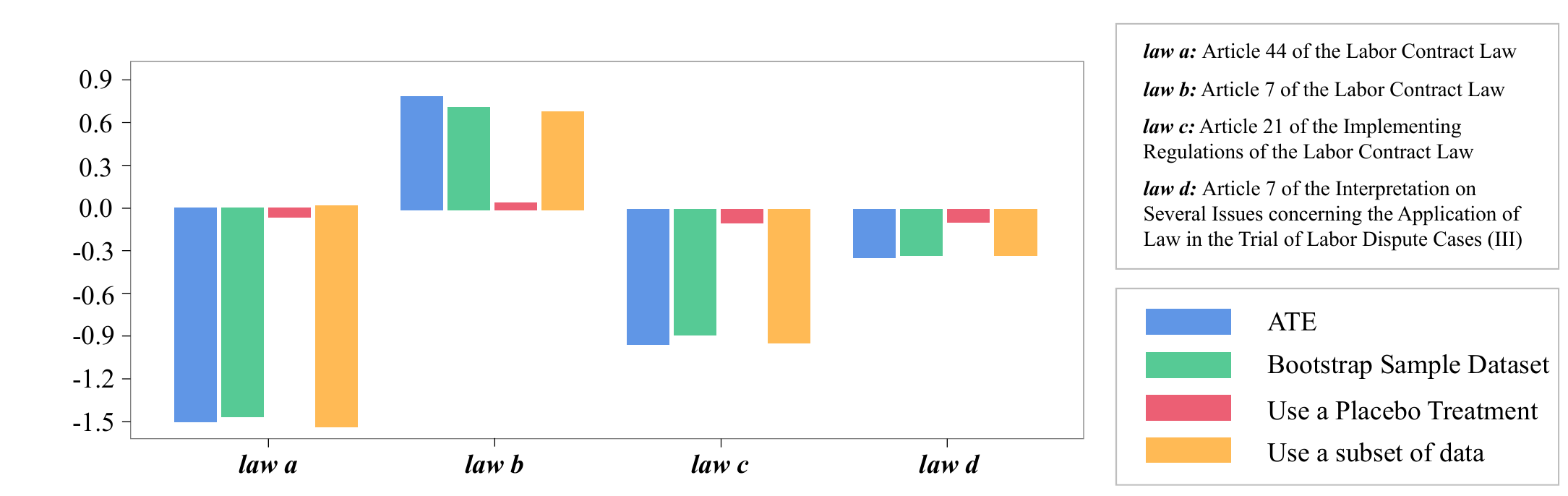}
\caption{Test results using three counterfactual methods.} \label{figure}
\end{figure}

To make our inference results more reliable, we conduct sensitivity tests to analyze the robustness of the results. Specifically, we use the counterfactual sample provided by Dowhy to generate counterfactual samples, and we use three counterfactual methods for testing. 1) Bootstrap Sample Dataset: Replacing a given dataset with bootstrap samples from the same dataset, it ideally does not show significant changes in causal effects. 2) Placebo Treatment: The real intervention variables were replaced with independent random variables, and the new ATE should be zero for the significant causal relationship that should not be exhibited between the variables in this condition. 3) Subset of Data: Replace the given dataset with a subset of data from the same dataset, ideally, the new ATE should remain the same as the previous one. The results of our sensitivity test are shown in Figure \ref{figure}. This demonstrates the significant robustness of our inference results. In addition, our inference results are agreed upon by legal experts.
\vspace{-4ex}

\subsection{Effect of Causal Smoothing}

\begin{figure}[H]
	\centering  
	\subfigbottomskip=2pt 
	\subfigcapskip=5pt 
	\subfigure[Control Group]{
		\includegraphics[width=0.3\linewidth]{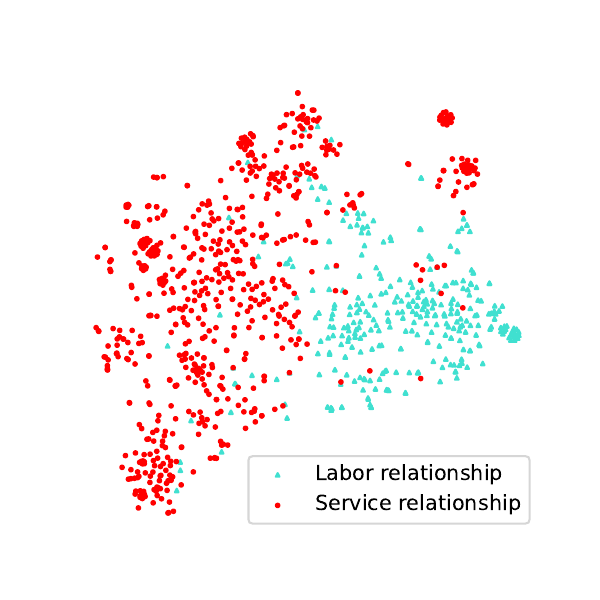}}
	\subfigure[Label Smoothing]{
		\includegraphics[width=0.3\linewidth]{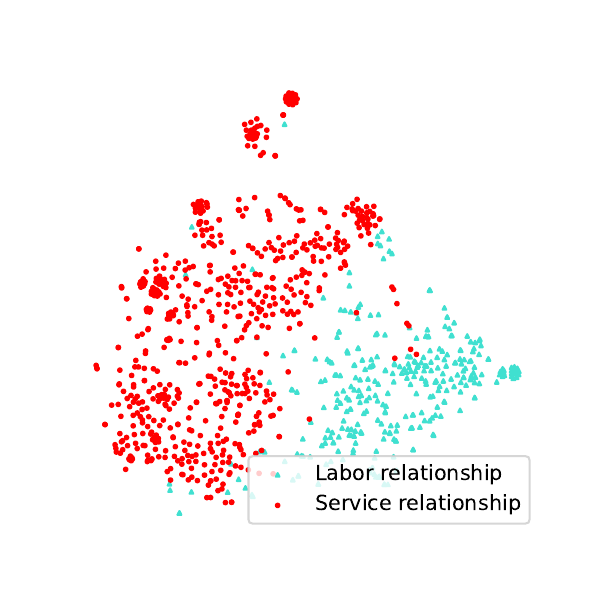}}
	\subfigure[Causal Smoothing]{
		\includegraphics[width=0.3\linewidth]{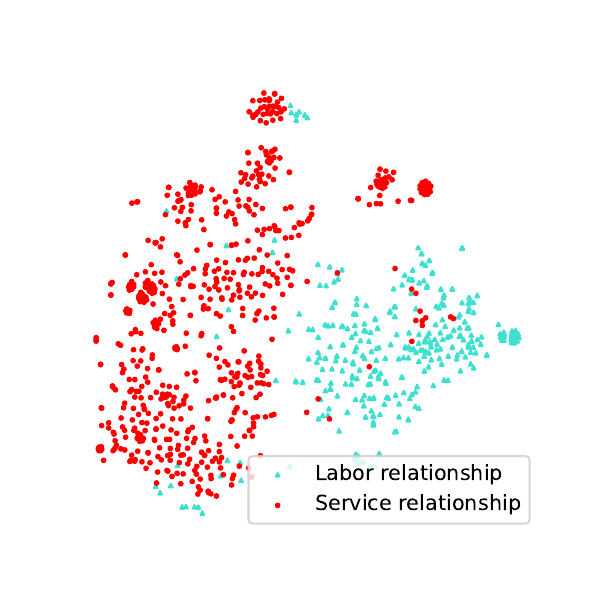}}
  \caption{The t-SNE plots of feature representations.} \label{smooth}
\end{figure}

  FAIR injects the causal knowledge obtained by the inference into the neural network model by Causal Smoothing and achieves significant improvement in downstream tasks. We believe this is because Causal Smoothing is closer to real judgment scenarios than Label Smoothing controlled by the hyperparameter $\epsilon$, which can simulate the judge's thoughts when deciding. For difficult judgments, Causal Smoothing makes its Soft Label closer to the judge's critical value, and the model is not overconfident in its prediction, which enhances the generalization of the model. To verify our conjecture, we select the output of the last hidden layer of the Lawformer model in the LJP experiment, downscale it using t-SNE~\cite{van2008visualizing}, and projected it onto a two-dimensional plane, and Figure \ref{smooth} shows our results. We can find that Label Smoothing reduces the intra-class distance to some extent compared to the Hard Label case, while Causal Smoothing shows the superiority of Causal Smoothing as the intra-class distance is more compact and the inter-class distance is pulled apart compared to the Label Smoothing case.
 \vspace{-6ex}

\section{Limitations of Large Language Model}
With the development of LLM, we become interested in the generalization capabilities of LLM in specific domains. In this section, we explore the capabilities of ChatGPT\footnote[2]{\url{https://openai.com/blog/chatgpt}} for legal intelligence downstream tasks and discuss its limitations. Our experiments have shown that improving the generalization ability of the LLM through the injection of causal knowledge can be achieved.

\subsubsection{Knowledge Utilization Capability} 
In the experiments shown in Table \ref{result}, we conducted LJP experiments with the legal scenario of over-age labors issues and compared it to our supervised training model. The results reveal that ChatGPT performs poorly without prior knowledge, with F1 values and accuracy both around 50. It suggests that ChatGPT struggles to utilize the knowledge of the over-age labors issues during training, and we believe this is because the LJP task is too challenging for a model trained without labels. Furthermore, it is difficult for the model to establish correlations between input facts and the laws learned during training without any additional cues. To investigate further, we incorporated judgment-related laws as prompts in the dialogue. Specifically, we add "you should pay attention to the use of the law x" (x=a, b, c, d) as prior knowledge in the prompt. The greater the influence of x in our inference results, the higher the level of the prompt. We find that ChatGPT's performance improved with different levels of prompts, but it still differed significantly from our supervised model. This indicates that ChatGPT can utilize input knowledge to some degree, but the prompt's design limits its utilization. Therefore, for legal intelligence downstream tasks, we require a supervised model tailored to the legal domain.

\subsubsection{Reasoning Ability} 
To utilize LLM in legal practice, we require the model's decisions to be highly interpretable. Thus, we conduct a fine-grained label extraction experiment to evaluate ChatGPT's reasoning ability in the legal domain.  We select six challenging labels from a dataset finely labeled by legal experts with the information shown in Table \ref{tab:array1}. This dataset will be publicly available soon. We take the original judgments as input and adjust the prompts several times to obtain the best performance, and Table \ref{tab:array2} shows the results of our experiments. We can find that ChatGPT can extract better for labels that may be given directly, such as C, while ChatGPT can barely extract effectively for labels that require inference from contextual descriptions to be obtained, such as F. Our experimental results show that ChatGPT still suffers from serious deficiencies in its inference ability in the legal domain, which blocks the application of LLM in the legal domain, and we hope that subsequent work can improve this.
\vspace{-6ex}
\begin{table}
    \centering
    \begin{minipage}[t]{0.53\textwidth}
        
        \caption{The challenging labels we select}    \label{tab:array1}
        \vspace{2ex}
        \begin{CJK*}{UTF8}{gbsn}
        \newcommand{\greycell}{\cellcolor[rgb]{0.9,0.9,0.9}}
        \resizebox*{\textwidth}{!}{
        \begin{tabular}{ll}
            \hline
            A Labor gender 劳动者性别 \\ 
            \multicolumn{2}{l}{ B \makecell[l]{When do labors to reach the mandatory age for retirement \\ 劳动者何时达到法定退休年龄 }  }\\ 
            \multicolumn{2}{l}{C Whether have  a written contract 有无书面合同  }\\
            \multicolumn{2}{l}{ D \makecell[l]{Whether enjoy the benefits of the old-age insurance \\ 有无享受养老保险待遇 } } \\ 
            \multicolumn{2}{l}{ E Kind of old-age insurance 养老保险待遇类型 }\\ 
            \multicolumn{2}{l}{ F \makecell[l]{Whether recognized of the basic old-age insurance \\ 是否认定为基本养老保险待遇 }  }\\
             \hline
        \end{tabular}
        }

        \end{CJK*}
    \end{minipage}
    \begin{minipage}[t]{0.46\textwidth}
\caption{Results on ChatGPT}\label{tab:array2}
\vspace{2ex}
    \resizebox*{\textwidth}{!}{
    \setlength{\tabcolsep}{2ex}{
        \begin{tabular}{ccccc}
   \toprule
   \textbf{Label} & \textbf{Acc} & \textbf{P} & \textbf{R} & \textbf{F1}\\
   \midrule
A & 56.25 & 56.40 & 56.73 & 55.77\\
B & 63.05 & 63.14 & 63.32 & 62.95\\
C & 70.84 & 72.40 & 72.06 & 70.82\\
D & 62.41 & 58.95 & 60.46 & 58.75\\
E & 61.53 & 38.25 & 42.20 & 35.07\\
F & 50.61 & 42.59 & 45.93 & 40.78\\
   \bottomrule
\end{tabular}
}}
    \end{minipage}

\end{table}
\vspace{-6ex}

\section{Conclusion}
We propose FAIR, a causal framework for accurately inferring judgment reversals, and we also introduce Causal Smoothing, a technique for incorporating causal knowledge into neural networks. In the context of predicting labor relationships of over-age labors, we demonstrate how the inferred causal effects enhance the model's performance. Our analysis examines the inferred outcomes' quality and sheds light on Causal Smoothing's role. Moreover, we undertake various tasks to evaluate large language models' capabilities in the legal domain. While acknowledging that LLM is not yet adequate for legal intelligence and cannot replace traditional supervised models. However, mining and injecting causal relationships can effectively enhance the generalization ability of the model, and improve the accuracy and fairness of legal result prediction.

\section{Ethics Statement}
We utilize a dataset sample sourced from publicly available judgment documents on the China Judgment Network, which is a platform that complies with relevant legal and regulatory requirements and authorizes the use of documents for research purposes. Our objective is to support legal services through FAIR principles and aid judges in their decision-making process rather than replace them. However, crucial information pertaining to over-age labors is often absent or ambiguous due to privacy concerns. This can result in the dataset being incomplete, potentially impacting the final analysis results. In certain cases, our model may generate erroneous judgments; hence users must exercise caution when interpreting the model's inference results. Nevertheless, on the whole, our model can assist judges in identifying pertinent legal articles and aid in ensuring judicial consistency throughout China.

%
%
%

\bibliographystyle{splncs04}
\bibliography{refbib}
%


\end{document}